\documentclass{mva_style}
\usepackage{graphicx}
\usepackage{color}
\usepackage{bm}
\usepackage{arydshln}
\usepackage[fleqn]{amsmath}

\finalcopy 

\begin{document}
\title{Enhancing Reliability of Medical Image Diagnosis\\ through Top-rank Learning with Rejection Module}

\author{
  Xiaotong Ji\\
  Kyushu University\\
  Fukuoka, Japan\\
  \and
  Ryoma Bise\\
  Kyushu University\\
  Fukuoka, Japan\\
  \and
  Seiichi Uchida\\
  Kyushu University\\
  Fukuoka, Japan\\
}

\maketitle

\section*{\centering Abstract}
\textit{
In medical image processing, accurate diagnosis is of paramount importance. Leveraging machine learning techniques, particularly top-rank learning, shows significant promise by focusing on the most crucial instances. However, challenges arise from noisy labels and class-ambiguous instances, which can severely hinder the top-rank objective, as they may be erroneously placed among the top-ranked instances. To address these, we propose a novel approach that enhances top-rank learning by integrating a rejection module. Co-optimized with the top-rank loss, this module identifies and mitigates the impact of outliers that hinder training effectiveness. The rejection module functions as an additional branch, assessing instances based on a rejection function that measures their deviation from the norm. Through experimental validation on a medical dataset, our methodology demonstrates its efficacy in detecting and mitigating outliers, improving the reliability and accuracy of medical image diagnoses.\footnote{The original paper has been accepted in MVA2025. ©2025 IEICE. Personal use of this material is permitted. Permission from IEICE must be obtained for all other uses.}
}

\section{Introduction}
\label{sec:intro}

The primary objective of medical image processing is to ensure reliable predictions, particularly by maximizing the detection of all relevant targets, such as pathological conditions or diseases \cite{bise2019cvpr}.
To optimize this, top-rank learning~\cite{li2014top, boyd2012accuracy} has emerged as a promising approach. It focuses on maximizing the number of reliable positive samples at the top of the ranking, which reduces the burden on healthcare practitioners by minimizing the need for manual labor in identifying obvious cases.
In essence, top-rank learning learns a ranking function that maximizes the number of "absolute positives," i.e., positive samples ranked higher than the top-ranked negatives. If the test distribution is similar to the training distribution, the model's performance on the test data should closely resemble its performance during training (as depicted in Fig.~\ref{fig:toprank_w_wo_outliers} (a) and (b)).

\setlength{\abovecaptionskip}{0cm}
\begin{figure}[t!]
\begin{center}
\setlength\abovecaptionskip{8.pt}
\includegraphics[width=\columnwidth]{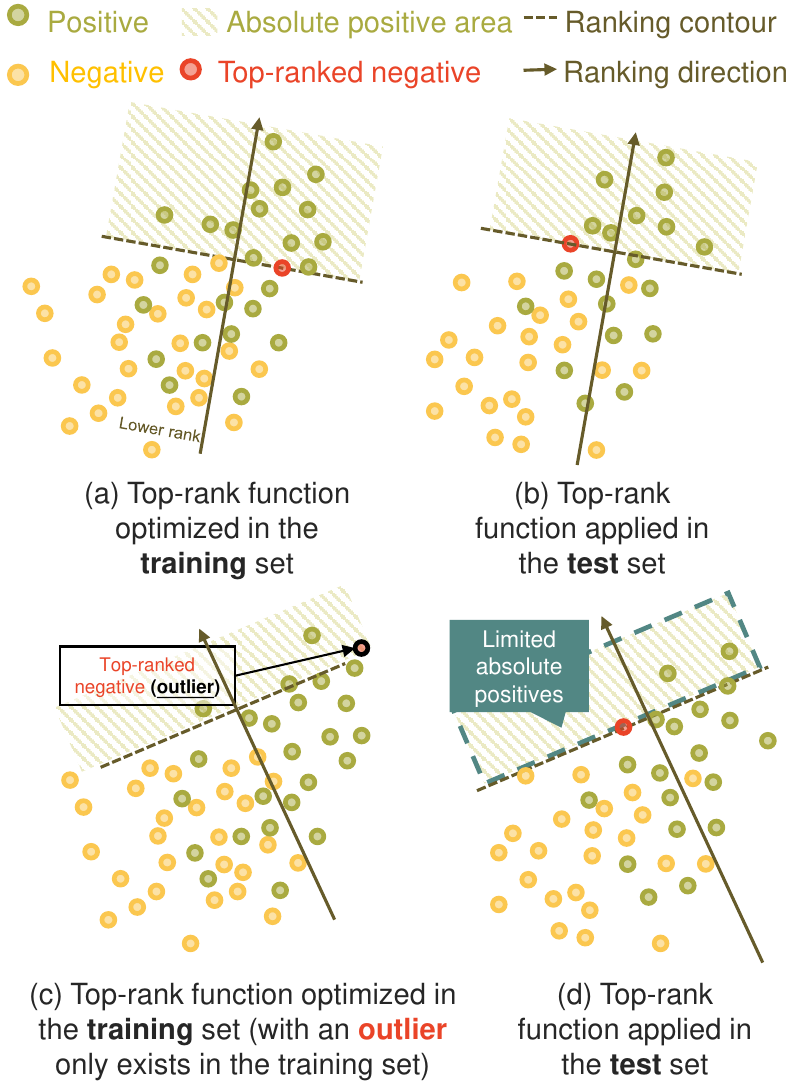}
\caption{
This figure illustrates how the presence of outliers during training can cause the top-ranked negatives to dominate the ranking surface, leading to a failure in identifying absolute positives at test time.
(a), (b) the training and test stages without outliers, respectively, and (c), (d) the same stages in the presence of outliers.
}
\label{fig:toprank_w_wo_outliers}
\vspace{-10pt}
\end{center}
\vspace{-10pt}
\end{figure}

However, outliers, particularly in the training data, pose a significant challenge to top-rank learning~\cite{zheng2021rob}. These outliers, which are placed at the top of the ranking, can disrupt the model's ability to learn and correctly identify positive samples at the forefront, as shown in Fig.~\ref{fig:toprank_w_wo_outliers} (c) and (d). When similar outliers do not exist in the test data, the top-rank function will struggle to identify sufficient absolute positives.
In medical image processing, intrinsic outliers often arise from data sourced from different medical institutions or variations in clinicians' judgment criteria.

Thus, the need to mitigate the impact of outliers on top-rank learning models becomes paramount to ensure generalizability.
Past research in medical image processing has attempted to tackle outlier detection through various methods, such as using Auto-Encoders~\cite{cowton2018combined,sato2018primitive}, Generative Adversarial Networks~\cite{schlegl2019f}, and Recurrent Neural Networks~\cite{fernando2020neural}. While these methods have shown promise, they often fall short when it comes to seamlessly integrating with predictive model optimization processes. 
In particular, joint optimization efforts to minimize outliers' impact during training while refining predictive models remain elusive.

To address the challenge posed by outliers, we propose a method that combines top-rank learning with a rejection function, optimized simultaneously. This approach weakens the influence of samples in the training data that hinder the top-rank learning function, resulting in a more generalizable ranking model for test data, where no rejection is needed during testing.
Importantly, this is not just a traditional ranking method combined with out-of-distribution (OOD) detection; rather, it selectively weakens outliers that affect ranking during the training process.

\section{Related Work}
\noindent
{\bf Top-rank learning:}
\label{subsec:TRL}
Top-rank learning aims to augment the identification of positively classified samples with an exceptionally high confidence level, delineating them as absolute positives and ensuring their precedence over any negative instances~\cite{frery2017efficient,li2014top,boyd2012accuracy}. 
By emphasizing the reliability of predictions, top-rank learning is especially suited for applications that require high confidence.
In the context of the burgeoning interest in representational learning, Zheng et al.~\cite{zheng2021top} have demonstrated the applicability of top-ranking learning with neural networks. They leverage $p$-norm push minimization techniques~\cite{rudin2009p} to regulate ranking concentration and mitigate potential overfitting issues.


\vspace{0.5\baselineskip}
\noindent
{\bf Outlier detection:}
Due to the significant ramifications of outliers, the field of medical image processing has long been dedicated to outlier detection~\cite{maniruzzaman2018accurate}. Deep learning-based methods, in particular, have shown superior computational efficiency, performance, and scalability~\cite{mishra2023dual, mehta2022out}.
Zheng et al.~\cite{zheng2021rob} introduce Local Outlier Factor (LOF)~\cite{breunig2000lof} into top-rank learning to identify anomalous data points by measuring their local deviation with respect to neighboring points. This approach addresses the issue of outliers within the training dataset. These methods depend on the outlier detection performances due to independent optimization.

Learning with Rejection~\cite{bartlett2008classification,cortes2016learning} has been proposed to counteract the overconfidence of neural networks in handling outlier data~\cite{nguyen2015deep}. This approach optimizes classification and rejection functions within a single model, thereby mitigating the impact of outliers. Geifman et al. demonstrated the efficiency of an end-to-end classification model that incorporates a co-optimized rejection module designed to discard samples that hinder the classification function's optimization~\cite{geifman2019selectivenet}. In this paper, we combine top-rank learning with a rejection function, optimizing both simultaneously.

\section{Top-rank Learning with Rejection}

\subsection{Top-Rank Learning Paradigm}

Top-rank learning is a specialized ranking paradigm that optimizes the arrangement of positive samples relative to negative samples, aiming to maximize the count of positive samples ranked above all negative samples, referred to as `absolute positives.'' These absolute positives represent highly reliable positive samples, surpassing even the top-ranked negative,'' denoted as $\max_{1\leq j\leq n}t(\boldsymbol{x}_j^-)$. 

Here, $\boldsymbol{x}_i^+$ and $\boldsymbol{x}_i^-$ represent positive and negative samples respectively, with $m$ and $n$ denoting their quantities. The function $t(\boldsymbol{x})$ signifies the top-rank function, yielding the ranking scores of input samples. 
The optimization function of top-rank learning is formulated as follows with $I(\cdot)$ denotes an indicator function:

\setlength{\abovedisplayskip}{4pt}
\setlength{\belowdisplayskip}{4pt}

\begin{align}
\label{eq:posatop}
    \mathrm{pos@top}=\frac{1}{m}\sum_{i=1}^m 
    I\left(t(\boldsymbol{x}_i^+) > 
    \max_{1\leq j\leq n}t(\boldsymbol{x}_j^-)
    \right).
\end{align}

\subsubsection{Integration with Neural Network (NN) Architecture}

To incorporate this top-rank learning paradigm into a neural network architecture, a p-norm relaxation can replace the max function to achieve a `milder'' top-rank optimization, thereby addressing the issue of overfitting. The resulting top-rank loss function, denoted as $\mathcal{L}_{\mathrm{Top}}$, is defined as:

\begin{align}
\label{eq:toprankloss}
    \mathcal{L}_{\mathrm{Top}}=
    \frac{1}{m}\sum_{i=1}^m\left(
    \sum_{j=1}^n\left(
    l(t(\boldsymbol{x}_i^+) - 
     t(\boldsymbol{x}_j^-)
    )\right)^p\right)^{\frac{1}{p}},
\end{align}
where $l(z) = \log(1+e^{-z})$ represents a surrogate loss function. 
The parameter $p$ dictates the degree of relaxation, where larger values of $p$ result in the p-normed function approximating the original function with a max function more closely. 

\setlength{\abovecaptionskip}{0cm}
\begin{figure}[t!]
\begin{center}
\setlength\abovecaptionskip{8.pt}
\includegraphics[width=\columnwidth]{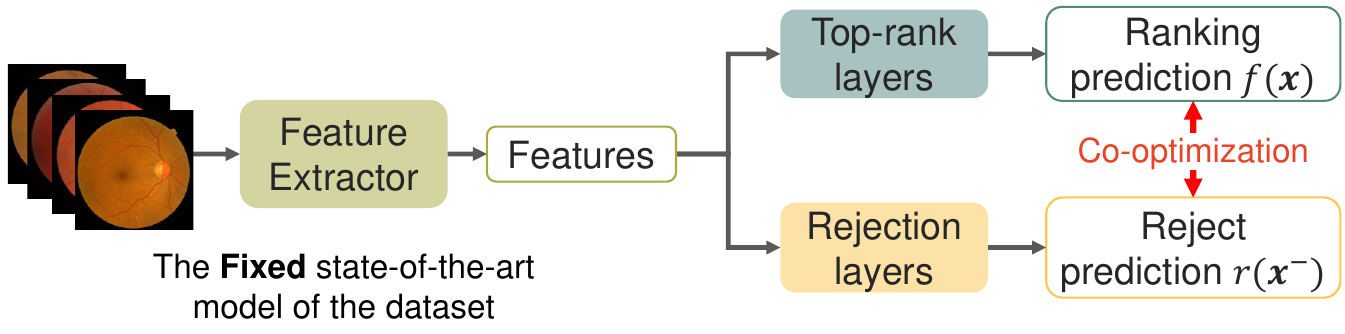}
\caption{Model structure of top-rank learning with rejection module.}
\label{fig:Model_structure}
\vspace{-10pt}
\end{center}
\end{figure}

\subsection{NN-based Top-rank Learning with Rejection}

To mitigate the detrimental impact of outliers on top-rank learning during training and ensure more generalized training, we propose utilizing a rejection module optimized alongside the top rank. The formulation is as follows:

\begin{align}
\label{eq:toprank+rej_loss}
\hspace{-3mm}\mathcal{L}_{\mathrm{TopRej}} & =
    \frac{1}{m}\sum_{i=1}^m\left(
    \sum_{j=1}^n\left(
    l(t(\boldsymbol{x}_i^+) - 
     t(\boldsymbol{x}_j^-)
    )r(\boldsymbol{x}_j^-)\right)^p
    \right)^{\frac{1}{p}} \notag \\
    & \quad +
    \lambda\max(0, c-\frac{1}{n}\sum_{i=1}^nr(\boldsymbol{x}_j^-))^2.
\end{align}

Notably, $r(\boldsymbol{x}^-)$ is applied only during training and exclusively to negative samples. It assigns lower weights to potential outliers that may harm the generalization of the top-rank function, thereby reducing their influence during optimization. 
To avoid arbitrarily weakening too many samples, the average of $r(\boldsymbol{x}^-)$ is softly regulated by a max-margin penalty. This ensures that only a portion of negative instances—presumed outliers—are attenuated. 
The degree of attenuation is controlled by the parameter $c$. For example, when $c = 0.9$, at most 10\% of the samples can be fully suppressed during training.

\subsection{NN-based Top-rank Learning with Rejection and LOF}

We also introduce an optimization function for top-rank learning with a rejection module and LOF score weighting to compare their effects and observe potential collaboration. Referencing Zheng et al.~\cite{zheng2021rob}, the formula incorporates LOF values of negative samples to adjust their weighting, attenuating the influence of negative outliers, where the hyperparameter $d$ controls the extent of outlier weakening:

\begin{align}
\label{eq:lof}
    O_j^d=\left(
    \frac{1}{\max(\mathrm{LOF}_k(\boldsymbol{x}_j^-), 1)}
    \right)^d.
\end{align}

Thus, the loss for the top-rank learning model with LOF is obtained as:

\begin{align}
\label{eq:toprank+LOF_loss}
    \mathcal{L}_{\mathrm{TopLOF}}=
    \frac{1}{m}\sum_{i=1}^m\left(
    \sum_{j=1}^n\left(
    l(t(\boldsymbol{x}_i^+) - 
     t(\boldsymbol{x}_j^-)
    )O_j^d\right)^p
    \right)^{\frac{1}{p}},
\end{align}

\noindent and a top-rank learning model simultaneously equipped with both rejection and LOF is formulated as:
\begin{align}
\label{eq:toprank+LOF_loss}
\hspace{-7mm}\mathcal{L}_{\mathrm{TopRejLOF}} & =
    \frac{1}{m}\sum_{i=1}^m\left(
    \sum_{j=1}^n\left(
    l(t(\boldsymbol{x}_i^+) - 
     t(\boldsymbol{x}_j^-)
    )r(\boldsymbol{x}_j^-)O_j^d\right)^p
    \right)^{\frac{1}{p}} \notag \\
    & \quad +
    \lambda\max(0, c-\frac{1}{n}\sum_{i=1}^nr(\boldsymbol{x}_j^-))^2.
\end{align}

\setlength{\abovecaptionskip}{0cm}
\begin{figure}[t!]
\begin{center}
\setlength\abovecaptionskip{8.pt}
\includegraphics[width=0.8\columnwidth]{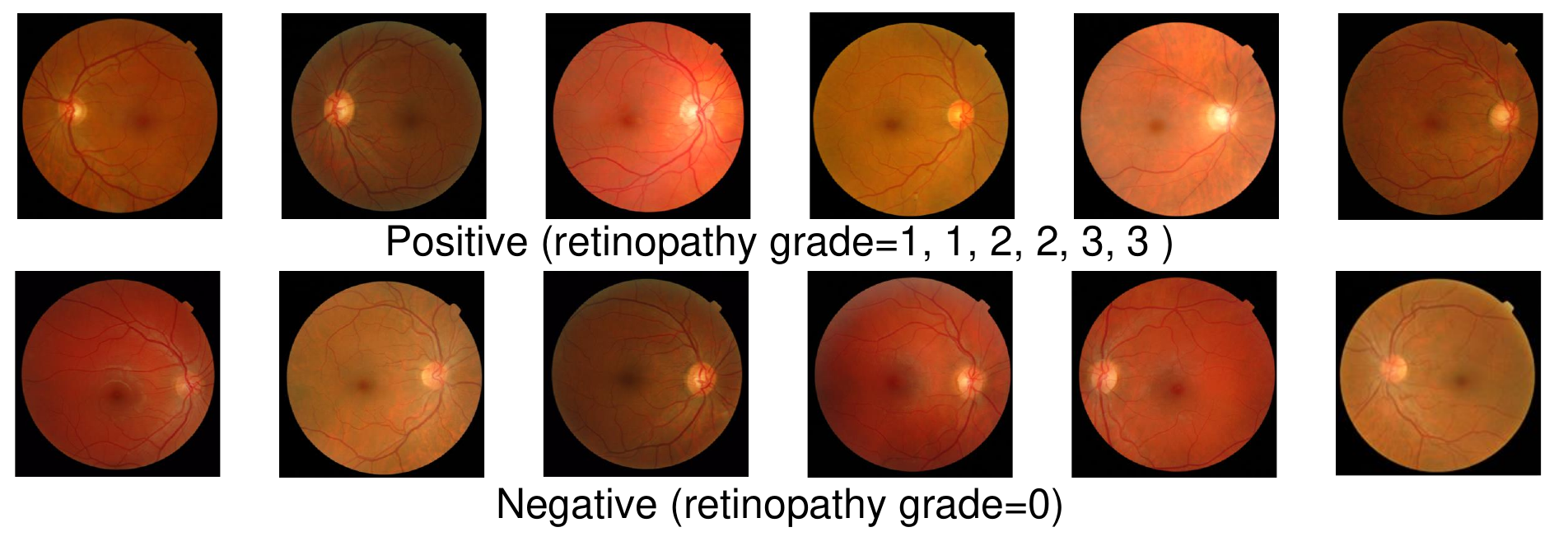}
\caption{Randomly selected samples from the Messidor dataset. This research considers any disease level a negative sample (bottom row) and the absence of disease a positive sample (top row).}
\label{fig:Messidor_examples}
\vspace{-14pt}
\end{center}
\end{figure}

\section{Experiment on Messidor Dataset}

\subsection{Dataset and Evaluation Metrics}

We practically demonstrate the proposed methodology through a diabetes retinopathy ranking experiment, utilizing the publicly available Messidor dataset~\cite{decenciere2014feedback}~\footnote{https://www.adcis.net/en/third-party/messidor/} which is designed for computer-assisted diagnoses. This dataset includes 1200 eye fundus color images from three ophthalmologic departments, annotated with retinopathy grades 0 (normal) to 3 (severe retinopathy). For this ranking research, we dichotomize these grades into 0 (normal) and 1 (retinopathy grade $\geq$ 1) to distinguish healthy individuals from patients specifically.
Examples of images from the Messidor dataset are illustrated in Fig.~\ref{fig:Messidor_examples}.

\setlength{\abovecaptionskip}{0cm}
\begin{figure}[t!]
\begin{center}
\setlength\abovecaptionskip{8.pt}
\includegraphics[width=0.55\columnwidth]{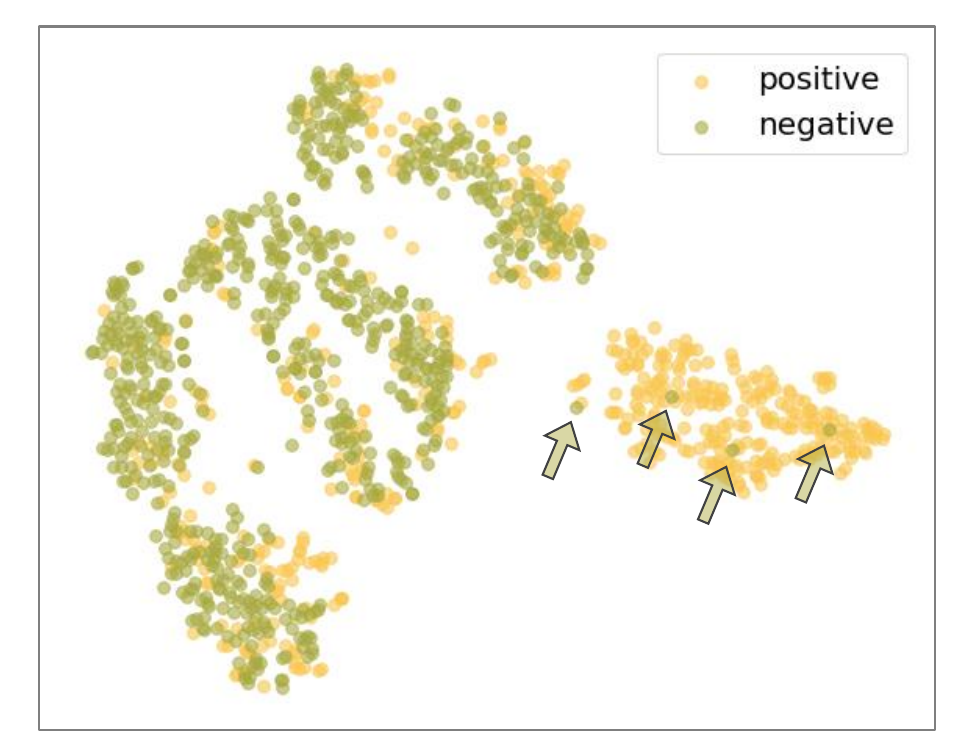}
\caption{The t-SNE visualization of features of the Messidor dataset (training and test sets) extracted from the trained CBNET. Arrows indicate negative outliers.}
\label{fig:Messidor_distribution}
\vspace{-14pt}
\end{center}
\end{figure}
We utilize a state-of-the-art classification model pre-trained on the Messidor dataset~\cite{he2020cabnet} as a fixed feature extractor to generate features for our proposed methodology. 
Both the training protocol and the data partition for ten-fold cross-validation align with this pre-training.
Moreover, as discussed in the introduction, the relatively higher negative outliers are especially detrimental to the ranking paradigm. The overall distribution of this dataset is depicted in Fig.~\ref{fig:Messidor_distribution}, where the outliers can be observed.
Among these metrics, pos@top, defined as the ratio of the number of absolute positives to the total number of positive samples, provides insight into the model's ability to detect relevant targets. In our case, where the targets are retinopathy samples, pos@top reflects how effectively the model identifies these samples.

\subsection{Proposed Models and Models for comparison}

The proposed methods include:
(1) a top-rank learning model with a rejection module (w. rej.), and
(2) a top-rank learning model with a rejection module and LOF (w. both).
Comparison methods consist of:
(3) a top-rank learning model~\cite{zheng2021top}, and
(4) a top-rank learning model with LOF~\cite{zheng2021rob} (w. LOF).
All models feature an NN structure, with the top-rank branch's structure being identical across all four models. The two proposed methods include an additional rejection branch.

Both the top-rank and rejection branches consist of two fully connected layers, with a ReLU activation function for the former and a sigmoid function for the latter. For each model, we vary the learning rate within [0.01, 0.1] with an interval of $0.01$, resulting in $11$ values. The parameter $p$ in Eqs.\ref{eq:toprankloss} and \ref{eq:toprank+rej_loss} takes values from ${16, 32, 64}$, while lambda is set to $32$ based on empirical observations. The rejection controlling parameter $c$ is set to $0.9$ based on Fig.\ref{fig:Messidor_distribution}. Parameter $d$ for the LOF in Eq.\ref{eq:lof} is set to $100$, and during top-rank module training, each step involves 5 positive samples and 45 negative samples to form pairs in the batches~\cite{zheng2021rob}.

\subsection{Result and Discussion}

\begin{table}[t]
\centering
\fontsize{8pt}{10pt}\selectfont
\caption{Comparison of four methods, with proposed methods in bold. The average of 10-fold cross-validation with variance is reported. The maximum result is in bold, and the second largest is underscored. "Top-rank" is omitted for the three methods below the first row.}\label{tab1}
\begin{tabular}{l|ccc}
\hline
Methods & pos@top($\uparrow$) & ROC-AUC($\uparrow$) & PR-AUC($\uparrow$) \\
\hline
Top-rank          & 0.481$\pm$0.185             & \textbf{0.907$\pm$0.033}  & 0.904$\pm$0.034\\
w. LOF            & 0.484$\pm$0.182             & \underline{0.904$\pm$0.037}           & 0.899$\pm$0.040\\
\textbf{w. rej.}      & \textbf{0.499$\pm$0.207}    & 0.902$\pm$0.275           & \underline{0.905$\pm$0.030}\\
\textbf{w. both} & \underline{0.497$\pm$0.190}             & 0.895$\pm$0.374           & \textbf{0.906$\pm$0.038}\\
\hline
\end{tabular}
\vspace{-12pt}
\end{table}

Table~\ref{tab1} presents the pos@top, ROC-AUC, and PR-AUC for four methods. It is evident that "Top-rank with Rejection" achieves the highest pos@top and the second-highest PR-AUC, highlighting its effectiveness in optimizing the top-ranked segment of predictions.

This is particularly valuable in clinical triage scenarios, where ranking absolute positives at the top is crucial. The rejection module contributes to better generalization on the test set by suppressing harmful outlier negatives during training.

\setlength{\abovecaptionskip}{0cm}
\begin{figure}[t!]
\begin{center}
\setlength\abovecaptionskip{8.pt}
\includegraphics[width=\columnwidth]{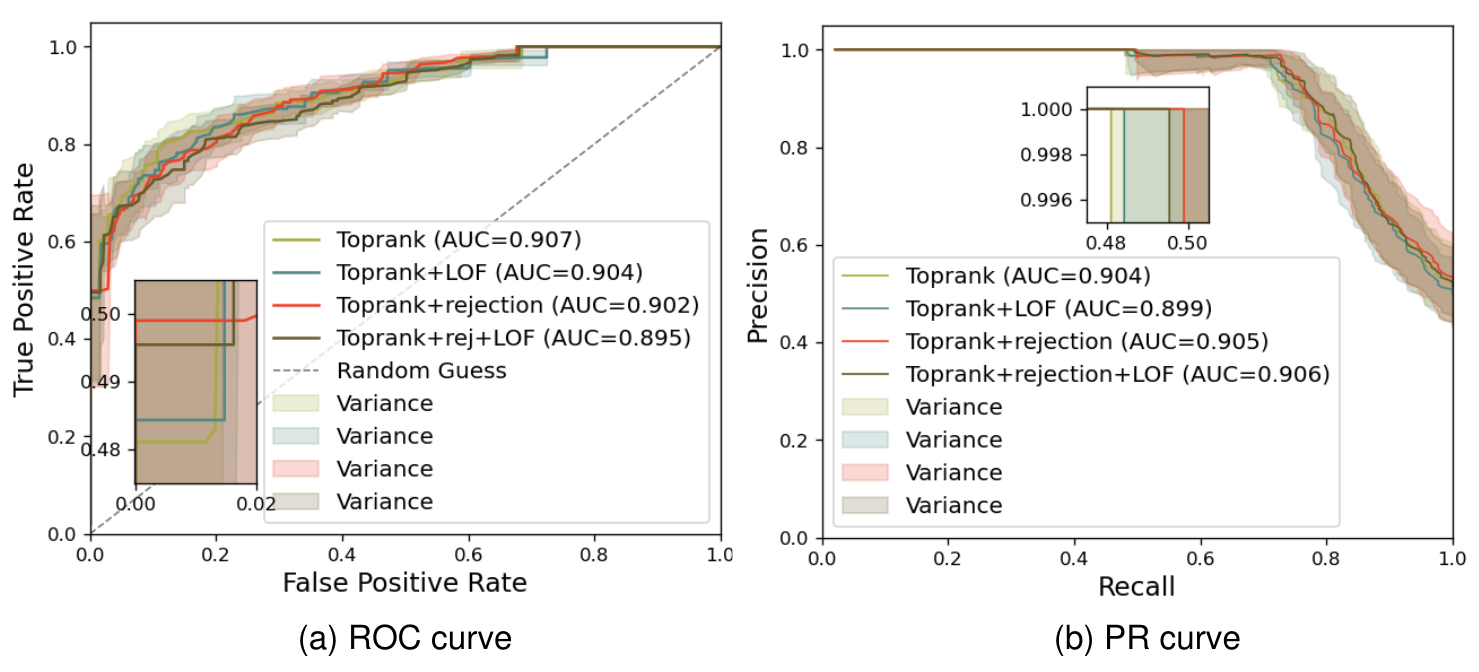}
\caption{(a)ROC curves and (b)PR curves for the four methods. pos@top equals the value of the true positive rate (TPR) at FPR=0. TPRs and precisions of the four methods can be clearly observed through the magnified partial curves.}
\label{fig:AUCs}
\vspace{-18pt}
\end{center}
\end{figure}

Similarly, `Top-rank with LOF with Rejection'' achieves the highest PR-AUC and the second-highest pos@top.
This outcome suggests that integrating rejection and LOF contributes to higher precision at the top ranking while maintaining a high pos@top. However, both methods exhibit inferior performance in ROC-AUC compared to models without rejection module, despite ROC-AUC not being the optimization objective of the top-rank learning method.

Figure~\ref{fig:AUCs} depicts the ROC and precision-recall curves of the four methods at various thresholds. The solid lines represent the average of 10-fold cross-validation, while the colored regions around them denote the variance. Notably, observing their initial inflection points provides insight into the magnitude of their respective pos@top values. 

\section{Conclusion}
We presented a top-rank learning framework with a rejection module to address the impact of outliers in medical image diagnosis. Such outliers may stem from variations in sampling protocols across institutions or inconsistencies in labeling criteria among annotators.
The rejection module is jointly optimized with the top-rank objective to detect and suppress potentially harmful negative samples, thereby improving model reliability and increasing the presence of absolute positives in the top-ranked outputs.
Experiments on a diabetic retinopathy dataset demonstrate the effectiveness of our method in mitigating the influence of outliers and enhancing diagnostic precision.

Future work includes extending this approach to multi-class and fine-grained classification tasks, as well as exploring adaptive rejection thresholds tailored to batch-level statistics.

\noindent{\bf Acknowledgments:}
This work was supported by SIP-JPJ012425, JSPS KAKENHI JP23K18509, JP24K03002, and JST JPMJAP2403.

\newpage

\bibliographystyle{ieeetran}
\bibliography{myrefs}

\end{document}